%% file: main.tex
\documentclass[manuscript,screen]{acmart}

\input{own_command}

\AtBeginDocument{%
  \providecommand\BibTeX{{%
    \normalfont B\kern-0.5em{\scshape i\kern-0.25em b}\kern-0.8em\TeX}}}

\setcopyright{acmlicensed}
\copyrightyear{2018}
\acmYear{2018}
\acmDOI{XXXXXXX.XXXXXXX}

\acmConference[Conference acronym 'XX]{Make sure to enter the correct
  conference title from your rights confirmation emai}{June 03--05,
  2018}{Woodstock, NY}
\acmISBN{978-1-4503-XXXX-X/18/06}

\begin{document}

\title
[
Benchmarking Large Language Models for Molecule Prediction Tasks
]
{
Benchmarking Large Language Models for Molecule Prediction Tasks
}

\author{Zhiqiang Zhong}
\affiliation{%
  \institution{Aarhus University}
  \country{Denmark}
}
\email{zzhong@cs.au.dk}

\author{Kuangyu Zhou}
\affiliation{%
  \institution{Microsoft}
  \country{China}
}
\email{luckyjoou@gmail.com}

\author{Davide Mottin}
\affiliation{%
  \institution{Aarhus University}
  \country{Denmark}
}
\email{davide@cs.au.dk}






\renewcommand{\shortauthors}{Zhong et al.}

\begin{abstract} 
\input{pages/abstract}
\end{abstract}




\maketitle

\section{Introduction} 
\label{sec:introduction}
\input{pages/introduction}

\section{Related Work} 
\label{sec:related_work}
\input{pages/related_work}

\section{Preliminary and Methodology} 
\label{sec:preliminary}
\input{pages/preliminary}


\section{Experiments} 
\label{sec:experiment}
\input{pages/experiment}


\section{Concluding Remarks} 
\label{sec:conclusion}
\input{pages/conclusion}

\begin{acks}
This work is supported by the Horizon Europe and Innovation Fund Denmark under the Eureka, Eurostar grant no E115712 - AAVanguard.
\end{acks}

\bibliographystyle{ACM-Reference-Format}
\bibliography{full_format_references}


\end{document}

%% file: own_command.tex
\usepackage{adjustbox}
\usepackage{makecell}
\usepackage{multirow}
\usepackage[ruled,vlined,linesnumbered]{algorithm2e}
\usepackage{enumitem}
\usepackage{amsmath,amsthm,amsfonts}
\usepackage{xcolor}
\usepackage{xspace}
\usepackage{mdframed}
\newmdtheoremenv[%
  backgroundcolor=gray!20,
  linecolor=red!60!black,
  linewidth=2pt,
  topline=false,
  rightline=false,
  skipabove=10pt,
  skipbelow=10pt,
  leftline=false]{regbox}{Box}

\newcommand{\specialcell}[2][c]{%
    \begin{tabular}[#1]{@{}c@{}}#2\end{tabular}
}

\newcommand{\lipo}{\texttt{ogbg-mollipo}\xspace}
\newcommand{\freesolv}{\texttt{ogbg-molfreesolv}\xspace}
\newcommand{\esol}{\texttt{ogbg-molesol}\xspace}
\newcommand{\bbbp}{\texttt{ogbg-molbbbp}\xspace}
\newcommand{\bace}{\texttt{ogbg-molbace}\xspace}
\newcommand{\hiv}{\texttt{ogbg-molhiv}\xspace}
\newcommand{\IF}{\textbf{IF}\xspace}
\newcommand{\IFD}{\textbf{IFD}\xspace}
\newcommand{\IP}{\textbf{IP}\xspace}
\newcommand{\IPD}{\textbf{IPD}\xspace}
\newcommand{\IE}{\textbf{IE}\xspace}
\newcommand{\IED}{\textbf{IED}\xspace}
\newcommand{\FS}{\textbf{FS}\xspace}

\newcommand{\FSD}{\textbf{FSD}\xspace}

\newcommand{\SAIF}{\textbf{S-IF}\xspace}
\newcommand{\SAIFD}{\textbf{S-IFD}\xspace}

\newcommand{\SAIE}{\textbf{S-IE}\xspace}
\newcommand{\SAIED}{\textbf{S-IED}\xspace}
\newcommand{\SAFS}{\textbf{S-FS}\xspace}

\newcommand{\SMILES}{\textbf{S}\xspace}
\newcommand{\solo}{\textsc{Solo}\xspace}
\newcommand{\duo}{\textsc{Duo}\xspace}
\newcommand{\trio}{\textsc{Trio}\xspace}

\newcommand{\eg}{\emph{e.g.}}
\newcommand{\ie}{\emph{i.e.}}



%% file: pages/abstract.tex
Large Language Models (LLMs) stand at the forefront of a number of Natural Language Processing (NLP) tasks. 
Despite the widespread adoption of LLMs in NLP, much of their potential in broader fields remains largely unexplored, and significant limitations persist in their design and implementation. 
Notably, LLMs struggle with structured data, such as graphs, and often falter when tasked with answering domain-specific questions requiring deep expertise, such as those in biology and chemistry.
In this paper, we explore a fundamental question: \emph{Can LLMs effectively handle molecule prediction tasks?} 
Rather than pursuing top-tier performance, our goal is to assess how LLMs can contribute to diverse molecule tasks. 
We identify several classification and regression prediction tasks across six standard molecule datasets. 
Subsequently,  we carefully design a set of prompts to query LLMs on these tasks and compare their performance with existing Machine Learning (ML) models, which include text-based models and those specifically designed for analysing the geometric structure of molecules. 
Our investigation reveals several key insights: 
Firstly, LLMs generally lag behind ML models in achieving competitive performance on molecule tasks, particularly when compared to models adept at capturing the geometric structure of molecules, highlighting the constrained ability of LLMs to comprehend graph data. 
Secondly, LLMs show promise in enhancing the performance of ML models when used collaboratively. 
Lastly, we engage in a discourse regarding the challenges and promising avenues to harness LLMs for molecule prediction tasks. 
The code and models are available at \url{https://github.com/zhiqiangzhongddu/LLMaMol}.

%% file: pages/introduction.tex
In recent decades, Machine Learning (ML) models have become increasingly prevalent in various real-world applications~\cite{BDCIW18,DHB20,ZBM23}. 
Both academia and industry have invested significant efforts in enhancing ML efficacy, aiming towards the realisation of Artificial General Intelligence (AGI) \cite{BCEG23}.
The remarkable advancements in generative models, such as Large Language Models ~\cite{VSPUJGKP17,DCLT18,BMRS20,OWJA22,ZZLTW23}, have ushered in a transformative era in Natural Language Processing (NLP). 
LLMs demonstrate unparalleled proficiency in comprehending and producing human-like text, proving indispensable in diverse NLP tasks such as machine translation \cite{AMAV23}, commonsense reasoning \cite{KMSG23}, and coding tasks \cite{BCEG23}.
A recent breakthrough known as In-Context Learning (ICL)~\cite{LYFJ23}, has further enhanced the adaptability of LLMs by enabling them to acquire task-specific knowledge during inference, reducing the need for extensive fine-tuning~\cite{CTR20}.

While LLMs have showcased their effectiveness across an array of NLP applications, the full extent of their potential in broader fields remains largely unexplored~\cite{ZDLW24}. 
Notably, LLMs encounter challenges with structured data like graphs and often struggle with domain-specific inquiries, such as those in biology and chemistry~\cite{BMKG23,JSOS24}.
To fill the gap, this paper delves into an essential research question: \emph{Can LLMs effectively handle molecule prediction tasks?}

To answer this research question, this paper identifies different important tasks, including classification and regression prediction tasks, across six benchmark molecule datasets~\cite{WRFGGPLP18,HFZDRLCL20}, \eg, \bace, \bbbp, \hiv, \esol, \freesolv and \lipo. 
Take a molecule, as illustrated in Figure~\ref{fig:molecule_info}, as an example, it can be represented in different representations, including \emph{SMILES string}~\cite{W88} and \emph{geometric structure}~\cite{ZDLW24}. 
However, a notable limitation of the existing LLMs is their reliance on unstructured text, rendering them unable to incorporate essential geometric structures as input~\cite{LLWL23,GDL23}. 
To address this challenge, \citet{FHP23} propose encoding the graph structure into text descriptions. 
In this paper, depicted in Figure~\ref{fig:molecule_info}, 
we extend this method by encoding both the molecule's atom features and graph structure into textual \emph{descriptions}.
Subsequently, we carefully design a set of prompts to harness various capabilities (\eg, domain-expertise, ICL capability) of LLMs to generate responses for molecule tasks.
Then we evaluate these responses in terms of consistency and performance on downstream tasks and compare them with those generated by existing ML models designed for molecule prediction tasks~\cite{HR18,ZCZ20}. 

The outcomes of our study effectively answered the raised question. 
Firstly, LLMs demonstrate a shortfall in competitive performance compared to existing ML models, particularly those specifically designed to capture the geometric structure of molecules. 
While ICL techniques offer notable assistance in improving LLM performance, they still trail behind existing ML models, underscoring the limited capability of current LLMs in directly addressing molecule tasks.
Secondly, we delve into the potential of integrating LLM responses with existing ML models, observing significant enhancements in numerous scenarios. 
We posit that leveraging LLMs as augmenters of domain knowledge currently presents a more effective approach than tasking LLMs with directly answering molecule predictive tasks.
In the end, we deliver a series of insightful discussions about limitations and promising avenues of existing LLMs in molecule tasks. We hope this work could shed new insight into the interdisciplinary framework design of molecule tasks empowered by LLMs. 

The rest of this paper is organised as follows. 
We begin by briefly reviewing related work in Section~\ref{sec:related_work}.
Afterwards, in Section~\ref{sec:preliminary}, we introduce the preliminaries of this study and include methodologies for molecule prediction tasks.
Experimental results are shown in Section~\ref{sec:experiment}.
Finally, we discuss the limitations and future work and conclude the paper in Section~\ref{sec:conclusion}.

%% file: pages/related_work.tex
\textbf{Large Language Models}.
Traditional language models are typically trained on sequences of tokens, learning the likelihood of the next token dependent on the previous tokens~\cite{VSPUJGKP17}. 
Recently, \citet{BMRS20} demonstrated that increasing the size of language models and the amount of training data can result in new capabilities, such as zero-shot generalisation, where models can perform text-based tasks without specific task-oriented training data. 
Consequently, Large Language Models (LLMs), such as GPT-3~\cite{BMRS20}, GPT-4~\cite{openaigpt4}, Flan-T5~\cite{CHLZ22}, Galactica~\cite{TKCS22}, Llama~\cite{TMSA23} and Gemini~\cite{TABW23}, have experienced exponential growth in both size and capability in recent years~\cite{AAAA23}. 
A wide range of NLP applications have been reshaped by LLMs, including machine translation~\cite{AMAV23}, commonsense reasoning~\cite{KMSG23} and coding tasks~\cite{BCEG23}.
While the impressive performance and generalisation capabilities of language models have rendered them highly effective across various tasks~\cite{WTBR22}, they have also resulted in larger model parameters and increased computational costs for additional fine-tuning on new downstream tasks~\cite{HSWA21}. 
To address this challenge, recent research has introduced In-Context Learning (ICL), enabling LLMs to excel at new tasks by incorporating a few task samples directly into the prompt~\cite{LYFJ23}. 
Despite their widespread use in NLP, their potential in broader fields remains largely unexplored. 
Thus, we conduct a comprehensive empirical analysis to evaluate the capability of LLMs in molecular prediction tasks. 

\smallskip\noindent
\textbf{Large Language Models Evaluation.}
In recent years, due to LLMs' great performance in handling different applications such as general natural language tasks and domain-specific ones, the evaluation of LLMs has become a significant field of inquiry~\cite{CWWW23}. 
\citet{GFQW23} showed ChatGPT’s proficiency in time-series prediction; \citet{AMAV23} evaluate the performance of LLMs are machine translators, while technical aspects of GPT-4 were analysed in~\cite{openaigpt4}. 
In the context of mathematical problem-solving, \citet{FPGS23} have highlighted that LLMs encounter challenges with graduate-level problems, primarily due to difficulties in parsing complex syntax.
Specifically, in healthcare, the utility and safety of LLMs in clinical settings were explored~\cite{NKMCH23}; \citet{YGWAZ24} explores the applications of the GPTs in general bioinformatics research, including gene and protein named entities extraction, identifying potential coding regions, etc. 

\smallskip\noindent
\textbf{Large Language Models for Molecular Tasks.}
Recent efforts integrating LLMs with the field of molecules mainly focus on harnessing the LLM's text-processing capabilities. 
For instance, \citet{GNLG23} established a comprehensive benchmark comprising eight practical chemistry tasks, specifically designed to gauge the performance of LLMs, including GPT-4 and GPT-3.5, across each task. 
\citet{ZZM24} harness the LLMs as post-hoc correctors to improve the ML model's molecule property predictions after completing training. 
Additionally, \citet{QTYLL23} utilised LLMs to generate explanations for molecule SMILES strings to facilitate predictions, while \citet{JSOS24} fine-tuned a GPT-3 model for addressing new chemical inquiries. 
This paper presents a comprehensive to evaluate the LLM's capability to handle both text and graph-structure information of molecules for prediction tasks. 
Furthermore, we showcase a straightforward yet efficacious framework for integrating LLMs with existing ML models.

%% file: pages/preliminary.tex
\begin{figure}[!ht]
\centering
\includegraphics[width=.6\linewidth]{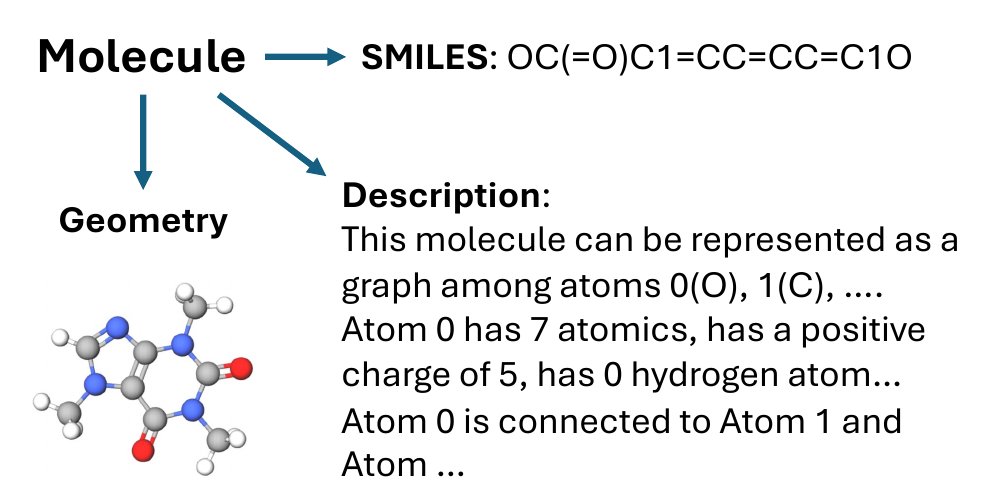}
\caption{
A molecule can be represented in different forms, \eg, SMILES string, text description and geometric structure. 
}
\label{fig:molecule_info}
\end{figure}

This paper aims to evaluate the capabilities of LLMs in handling challenging prediction tasks on structured molecule data within the field of biology.
For a given molecule, we can represent it using various formats such as \emph{SMILES (Simplified molecule Input Line Entry System) string}~\cite{W88} and \emph{geometric structures}~\cite{ZDLW24} (as shown in Figure~\ref{fig:molecule_info}). 
However, a notable limitation of existing LLMs is their reliance on unstructured text, rendering them unable to incorporate essential geometric structures as input~\cite{LLWL23,GDL23}. 
To overcome this limitation, \citet{FHP23} propose encoding the graph structure into text descriptions, and they have verified the effectiveness of such a design in various graph reasoning tasks. 
In this paper, as depicted in Figure~\ref{fig:molecule_info}, 
we extend this method by encoding both the molecule's atom features and graph structure into textual \emph{descriptions} since the molecule's atom features are important for different prediction tasks~\cite{HFRNDL21,ZBM23}. 

\subsection{Problem Setup}
\label{subsec:problem_setup}

\smallskip
\noindent
\textbf{Notion.}
Given a molecule, we formally represent it as $\mathcal{G} = (S, G, D)$, where $S$, and $G$ denote the \emph{SMILES string}, and \emph{geometric structure}.
$D$ indicates the generated atom feature and graph structure \emph{descriptions} of $\mathcal{G}$, as illustrated in Figure~\ref{fig:molecule_info}. 
$y \in \mathcal{Y}$ stands for the label for $\mathcal{G}$. 

\smallskip
\noindent
\textbf{Problem Setup.}
Given a set of molecules $\mathcal{M} = \{ \mathcal{G}_{1}, \mathcal{G}_{2}, \dots, \mathcal{G}_{m}\}$, where $\mathcal{M}_{\mathcal{T}} \subset \mathcal{M}$ contains molecules with known labels $y_{v}$ for all $\mathcal{G}_{v} \in \mathcal{M}_{\mathcal{T}}$. 
Our objective is to predict the unknown labels $y_{u}$ for all $\mathcal{G}_{u} \in \mathcal{M}_{test}$, where $ \mathcal{M}_{test} = \mathcal{M} \setminus \mathcal{M}_{\mathcal{T}}$. 
In addition, $\mathcal{M}_{\mathcal{T}}$ is split into two subsets: $\mathcal{M}_{train}$ and $\mathcal{M}_{val}$, where $\mathcal{M}_{train}$ is the training set and $\mathcal{M}_{val}$ works as the validation set. 
This separation allows us to fine-tune model parameters, mitigating overfitting, and validate the effectiveness of Machine Learning (ML) models before applying it to test dataset $\mathcal{M}{test}$.

\smallskip
\noindent
\textbf{ML Models for Molecule Prediction Tasks.}
The conventional approach to tackle molecule property prediction tasks is employing ML models. 
Take the supervised molecule property prediction task as an example. 
The goal is to learn a mapping function $f_{ML}: \mathcal{M} \to \hat{\mathcal{Y}}$, by minimising loss function value $\min_{\Theta} \sum_{i=1}^{n} \mathcal{L}(\hat{\mathcal{Y}}_{train}^{i}, \mathcal{Y}_{train}^{i})$, where $\Theta$ represents the set of trainable parameters of $f_{ML}$. 
Subsequently, $f_{ML}$ can be employed on test dataset $\mathcal{M}_{test}$ to generate predictions $\hat{\mathcal{Y}}_{test}$. 

\subsection{Prompt Engineering}
\label{subsec:prompt_engineering}

The goal in prompt engineering is to find the correct way to formulate a question $\mathcal{Q}$ in such a way that an LLM ($f_{LLM}$) will return the corresponding answer $A$ essentially represented as $A = f_{LLM}(\mathcal{Q})$. 
In this work, our goal is to provide the LLM with helpful and comprehensive knowledge regarding molecules so that it can make predictions on the test dataset. 
A variety of approaches exist for modifying the $f_{LLM}$ so that it could better perform downstream tasks such as fine-tuning~\cite{CTR20} and LoRA~\cite{HSWA21}. 
However, these methods typically require access to the internals of the model and heavy computation capability, which can limit their applicability in many real-world scenarios. 
In this work, we are instead interested in the case where $f_{LLM}$ and its parameters are fixed, and the system is available only for users in a black box setup where $f_{LLM}$ only consumes and produces text. 
We believe this setting to be particularly valuable as the number of proprietary models available and their hardware demands increase. 

\begin{regbox} \emph{\textbf{Input-Feature (\IF) Prompt Example}}
\label{box:if}
\begin{verbatim}
<Instruction>
You are an expert in biomedicine and chemistry, specializing in molecules.
</Instruction> 
<Question>
Provide detailed insights on the molecule with the SMILES string {SMILES STRING}. {DESCRIPTION}
I am particularly interested in understanding whether the molecule {TASK}.
</Question> 
\end{verbatim}
\end{regbox}

\begin{regbox} \emph{\textbf{Input-Prediction (\IP) Prompt Example}}
\label{box:ip}
\begin{verbatim}
<Instruction> 
You are an expert in biomedicine and chemistry, specializing in molecules.
</Instruction> 
<Question>
The SMILES string of molecule-{ID} is {SMILES STRING}. {DESCRIPTION} 
Predict whether molecule-{ID} {TASK}. 
Answer the question in the format: Prediction: <True or False>.
</Question> 
\end{verbatim}
\end{regbox}

\begin{regbox} \emph{\textbf{Input-Explanation (\IE) Prompt Example}}
\label{box:ie}
\begin{verbatim}
<Instruction>
You are an expert in biomedicine and chemistry, specializing in molecules.
</Instruction>
<Question>
The SMILES string of molecule-{ID} is {SMILES STRING}. {DESCRIPTION} 
Predict whether molecule-{ID} {TASK}. 
Answer the question in the format: Prediction: <True or False>; Explanation: <text>.
</Question> 
\end{verbatim}
\end{regbox}

\smallskip\noindent
\textbf{Zero-shot Prompting}.
To this end, the first set of prompts (\IF, \IP, \IE) simply provide the LLM with molecule SMILES string $S$ and descriptions $D$ and asks it to generate the desired output with a desired format without any prior training or knowledge on the task, as illustrated in Box~\ref{box:if}, Box~\ref{box:ip} and Box~\ref{box:ie}~\cite{BMRS20}. 
The only guidance we provide to the LLM is \emph{instruction}, which tells about a little background context. 
Particularly, \IF asks the LLM to provide meaningful insights that might be helpful for the prediction task~\cite{QTYLL23}. 
\IP only asks the LLM to provide predictions about the molecule's properties, while \IE further asks for explanations, which may require the LLM to clarify the thought process in explanation generation and provide helpful evidence to help users understand the given prediction. 
In addition, if we fill out the \emph{description} of \IF, \IP and \IE, which derives \IFD, \IPD and \IED prompts. 
\emph{description} provides more comprehensive information about the molecule graph's features and structure information, yet it introduces a significant number of tokens, which can affect the LLM's answer consistency and over the constraints of LLMs.
We will address corresponding details in Section~\ref{sec:experiment}. 

\begin{regbox} \emph{\textbf{Few-shot (\FS) Prompt Example}}
\label{box:fs}
\begin{verbatim}
<Instruction>
You are an expert in biomedicine and chemistry, specializing in molecules.
</Instruction>
<Knowledge>
<Question>
The SMILES string of molecule-{ID} is {SMILES STRING}. {DESCRIPTION} 
Predict whether molecule-{ID} {TASK}.
</Question>
<Response>
Molecule-{ID} {ANSWER}. 
//Example classification ANSWER is "can inhibit human BACE-1."
</Response>
...
</Knowledge>
<Question>
The SMILES string of molecule-{ID} is {SMILES STRING}. {DESCRIPTION} 
Predict whether molecule-{ID} {TASK}. 
Answer the question in the format: Prediction: <True or False>; Explanation: <text>.
</Question> 
\end{verbatim}
\end{regbox}

\smallskip\noindent
\textbf{Few-shot Prompting}.
The second set of prompts (\FS) that we propose provides the LLM with a small number of examples of the task, along with the desired outputs~\cite{BMRS20}.
The LLM then learns from these examples to perform the task on new inputs. 
This approach can be categorised as a simple In-Context Learning (ICL) technique, 
An example prompt template is shown in Box~\ref{box:fs}. 
\FS-$k$ indicates $k$ contextual knowledge instances are included in the prompt. 
In this work, we do not discuss the \FSD prompts since the generated descriptions have tons of tokens, which will easily go over the LLM's input constraints. 

We note there are also some popular recent ICL techniques, \eg, Chain-of-thought (CoT)~\cite{WWSBIXCLZ22}, Tree-of-thought (ToT)~\cite{YYZSGCN23}, Graph-of-thought (GoT)~\cite{BBKG23} and Retrieval Augmented Generation (RaG)~\cite{LPPP20}, which are theoretically available to support complicated tasks and include large knowledge context. 
However, our initial experiments showed that methods, \eg, CoT, ToT and GoT, perform much worse for molecule property prediction tasks due to the significant difficulties in designing proper chain thoughts without solid expertise.
RaG implementations that we tested are unstable and slow with query, and they fall short of the relatively simpler \FS's performance. 
We argue it is caused by the unqualified information retrieval system, and we will discuss it in the future work discussion section. 

\subsection{Predictive Models}
\label{subsec:predictive_models}

To generate predictions on target molecules $\mathcal{M}_{test}$, this section presents our predictive models, including Large Language Model (LLM)~\cite{ZZLTW23}, Language Model (LM)~\cite{HR18} and Graph Neural Network (GNN)~\cite{ZCZ20} based approaches, which can capture comprehensive molecule information. 
In the following section, we will discuss their details. 

\begin{figure*}[!ht]
\centering
\includegraphics[width=.8\linewidth]{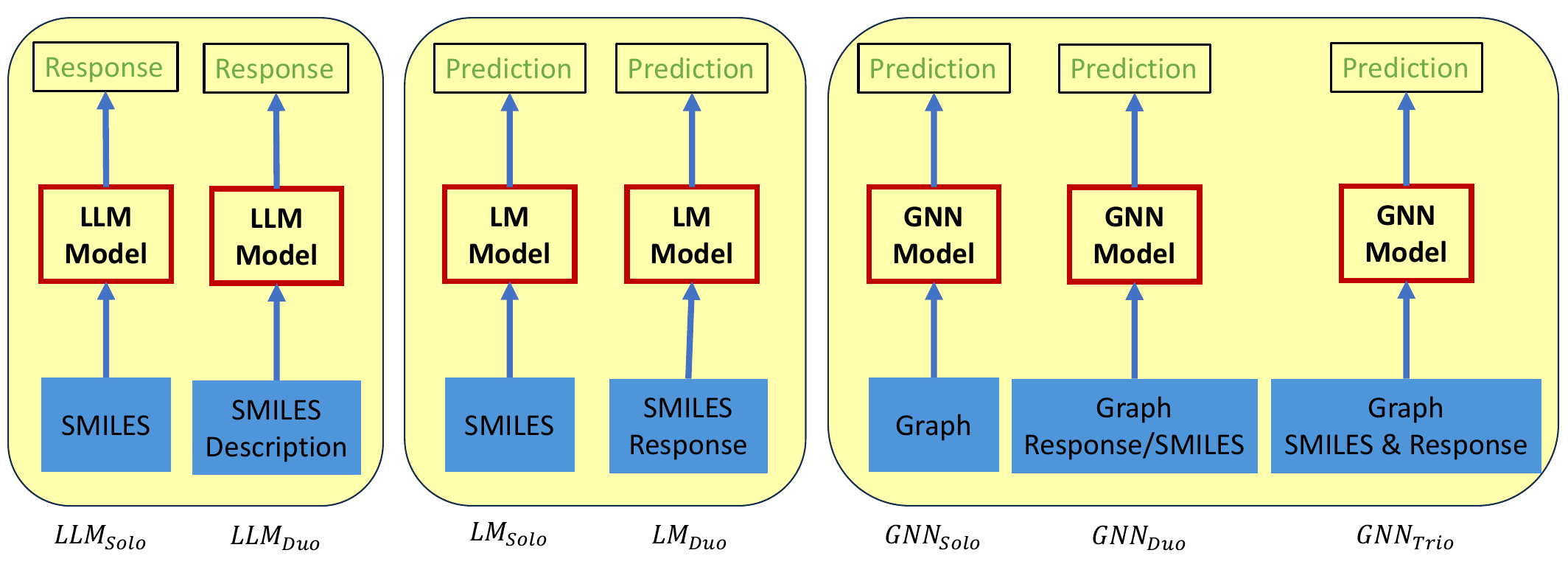}
\caption{
An overview of our pipelines for different predictive models. 
}
\label{fig:pipeline}
\end{figure*}

\smallskip\noindent
\textbf{LLM-based Approach.}
The LLM-based approaches take generated prompts, which follow the templates as discussed in Section~\ref{subsec:prompt_engineering}, as inputs, generating answers $R$ following the given format: $R = f_{LLM}(\mathcal{Q})$.
In addition, depending on the number of input information, we categorise them as LLM$_{\solo}$ and LLM$_{\duo}$ approaches, as shown in Figure~\ref{fig:pipeline} 
Particularly, LLM$_{\solo}$ takes queries based on \IF, \IP, \IE and \FS templates as input, LLM$_{\duo}$ takes queries based on \IFD, \IPD and \IED templates as input. 

\smallskip\noindent
\textbf{LM-based Approach.}
LMs can generate predictions based on available textual information~\cite{HR18}, \eg, SMILES string $S$, descriptions $D$ and the response $R$ provided by the LLM model. 
Because our empirical studies find that the performances of LM models utilising descriptions $D$ are not competitive compared with other settings. 
Therefore, as illustrated in Figure~\ref{fig:pipeline}, this work adopts two designs, \ie, only taking SMILES string as input (\textsc{LM}$_{\solo}$) and taking SMILES string and responses provided by the LLM as input (\textsc{LM}$_{\duo}$): $\hat{\mathcal{Y}} = f_{LM}(S, R)$~\cite{DCLT18}. 

\begin{figure*}[!ht]
\centering
\includegraphics[width=.7\linewidth]{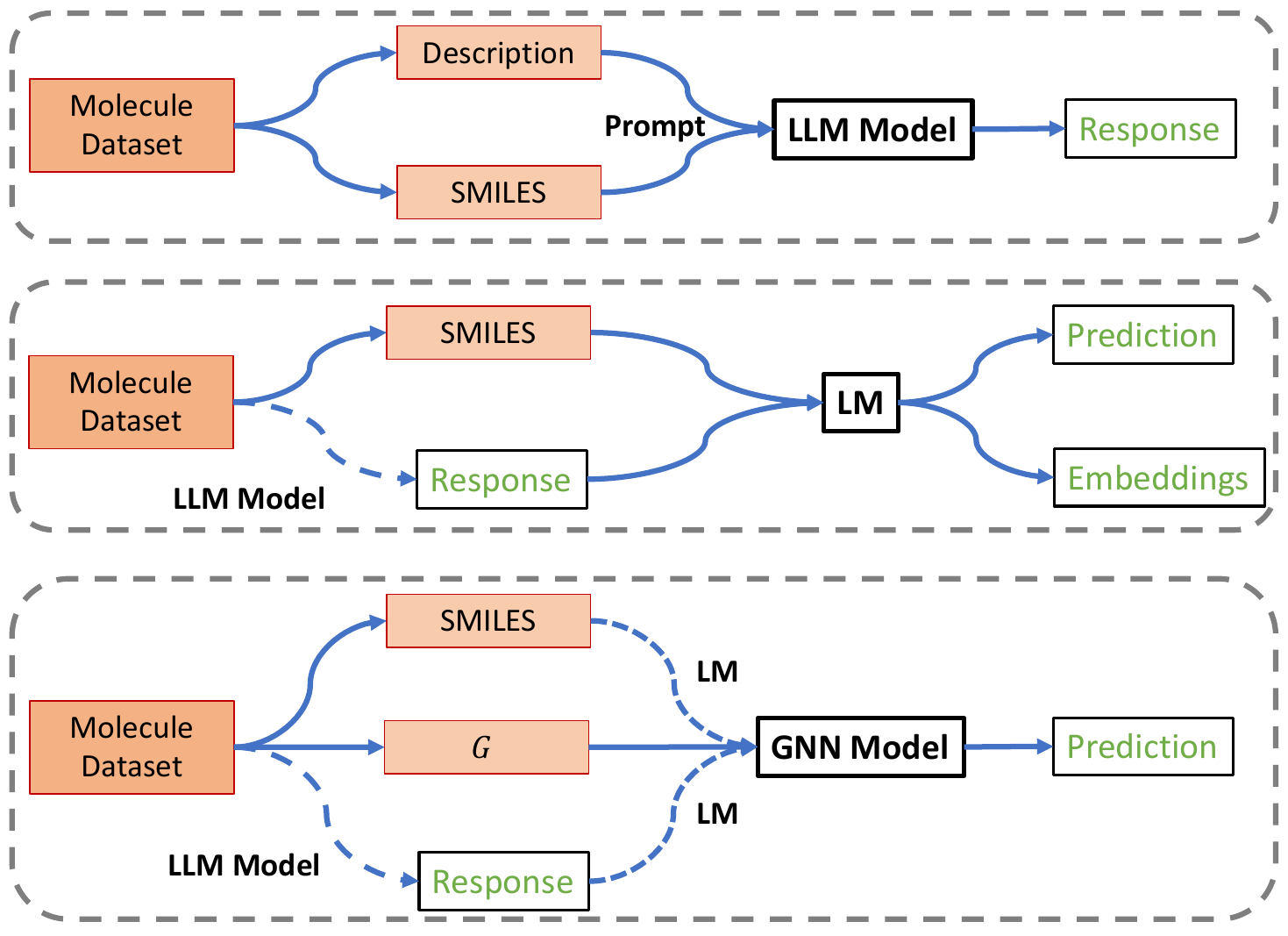}
\caption{
An illustration of our frameworks for different predictive models. 
}
\label{fig:framework}
\end{figure*}

\smallskip\noindent
\textbf{GNN-based Approach.}
GNN models are state-of-the-art approaches to molecule property prediction tasks since they are highly effective in capturing different essential geometric structure information $G$ of molecules. 
In addition, with the assistance of LMs, available textual information can also be converted into additional features ($X$) and fed into the GNN models afterwards: $\hat{\mathcal{Y}} = f_{GNN}(G, X)$. 
In particular, as shown in Figure~\ref{fig:framework}, benefits from the LM's flexibility in converting textual information into embeddings, which empowers GNN models with flexibility in incorporating information from different perspectives. 
This work adopts three design, \ie, GNN$_{\solo}$, GNN$_{\duo}$ and GNN$_{\trio}$, as illustrated in Figure~\ref{fig:pipeline}. 

%% file: pages/experiment.tex
This section presents our empirical studies and analysis in evaluating the effectiveness of LLMs in molecule prediction tasks. 
Our experimental analysis focuses on the challenging molecule graph property prediction tasks. 
We will present the experimental setup in Section~\ref{subsec:experiment_setup}, then demonstrate and discuss the experimental results in Section~\ref{subsec:performance}. 

\subsection{Experiment Setup}
\label{subsec:experiment_setup}

\input{tables/table-dataset}

\smallskip\noindent
\textbf{Dataset.}
We consider six benchmark molecule property prediction datasets that are common within ML research, including \bace, \bbbp, \hiv, \esol, \freesolv and \lipo. 
The collected datasets are summarised in Table~\ref{table:dataset}. 
Their detailed descriptions are as follow:
\begin{enumerate}
    \item \bace. The \bace dataset offers both quantitative ($\mathrm{IC}_{50}$) and qualitative (binary label) binding results for a series of inhibitors targeting human $\beta$-secretase 1 (BACE-1). 
    These data comprise experimental values sourced from scientific literature spanning the last decade, including instances accompanied by detailed crystal structures. 
    MoleculeNet~\cite{WRFGGPLP18} has amalgamated a repository of 1,522 compounds, incorporating their 2D structures alongside binary labels, formulated as a classification challenge.
    \item \bbbp. The Blood–Brain Barrier Penetration (BBBP) dataset originates from scientific investigations focused on modelling and forecasting barrier permeability. 
    The blood-brain barrier functions as a membrane that separates circulating blood from brain extracellular fluid. 
    It obstructs the majority of drugs, hormones, and neurotransmitters. Consequently, traversing this barrier has posed a persistent challenge in drug development aimed at the central nervous system. 
    This dataset encompasses binary labels assigned to over 2,039 compounds based on their permeability characteristics. Scaffold splitting is also advised for this clearly defined target.
    \item \hiv. The HIV dataset originated from the Drug Therapeutics Program (DTP) AIDS Antiviral Screen, which assessed the capacity to hinder HIV replication across 41,127 compounds. 
    Screening outcomes were categorised into three groups: confirmed inactive (CI), confirmed active (CA), and confirmed moderately active (CM). We subsequently amalgamated the latter two labels, transforming them into a classification task distinguishing between inactive (CI) and active (CA and CM) compounds. 
    Since our focus is on uncovering novel categories of HIV inhibitors, scaffold splitting is advised for this dataset.
    \item \esol. The ESOL dataset comprises water solubility data for 1,128 compounds. 
    It has been employed to train models aimed at estimating solubility directly from chemical structures encoded in SMILES strings. 
    Notably, these structures lack 3D coordinates, as solubility pertains to the molecule itself rather than its specific conformers.
    \item \freesolv. The Free Solvation Database (FreeSolv) offers both experimental and calculated hydration-free energies of small molecules in water. 
    A subset of these compounds is also utilised in the SAMPL blind prediction challenge. 
    The calculated values are obtained through alchemical free energy calculations employing molecular dynamics simulations. 
    While the experimental values are incorporated into the benchmark collection, the calculated values are utilised for comparison purposes.
    \item \lipo. Lipophilicity stands as a crucial characteristic of drug molecules, influencing membrane permeability and solubility alike. 
    Sourced from the ChEMBL database~\cite{MGBCDFMMMN19}, this dataset furnishes experimental findings on the octanol/water distribution coefficient (log D at pH 7.4) for 4200 compounds.
\end{enumerate}

\smallskip\noindent
\textbf{ML Models.}
To investigate the effectiveness of LLMs on molecule prediction tasks. 
We consider ML models of two different categories: (1) Language Model (LM) that only takes text information as inputs, \ie, DeBERTa~\cite{HGC21}. 
(2) Graph Neural Networks (GNNs) that capture the molecule's geometric structure information and other available features. 
We consider two classic GNN variants, \ie, GCN~\cite{KW17} and GIN~\cite{XHLJ19}. 
Their frameworks are illustrated in Figure~\ref{fig:framework}. 

\smallskip\noindent
\textbf{LLMs.}
In this work, we are interested in where the LLM's parameters are fixed, and the system is available for users in a black box setup where the LLM only consumes and produces text. 
We believe this setting to be particularly valuable as most users would practically have access to LLMs. 
In this case, we consider Llama-2-7b~\cite{TMSA23}, Llama-2-13b~\cite{TMSA23}, GPT-3.5 and GPT-4 \cite{AAAA23} as LLMs in this work, and GPT-3.5 is the major LLM for most experiments. 
The reasoning for this choice will be addressed in Section~\ref{subsec:performance}. 
All responses are obtained by calling their official APIs or their official implementation on \url{https://huggingface.co}. 
Because the generated \emph{descriptions} following \cite{FHP23} have tons of tokens, easily over the LLM's input token constraints, hence we do not include descriptions in the \FS prompt in this study. 

\smallskip\noindent
\textbf{Implementation.}
We implement ML predictive models following their available official implementations.
For instance, we adopt the available code of variant GNN models on the OGB benchmark leaderboards, \eg, GCN\footnote{\url{https://github.com/snap-stanford/ogb/tree/master/examples/graphproppred/mol}} and GIN\footnote{\url{https://github.com/snap-stanford/ogb/tree/master/examples/graphproppred/mol}}. 
About DeBERTa, we adopt its official implementation~\footnote{\url{https://huggingface.co/microsoft/deberta-v3-base}} and incorporate it within our pipeline. 
For the LLMs, we simply call the API provided by OpenAI or the official implementation with default hyper-parameter settings. 
We empirically tried with some combinations of recommended important hyper-parameters, \eg, temperature and top\_P, yet did not observe significant improvement. 

\begin{figure*}[!ht]
\centering
\includegraphics[width=.8\linewidth]{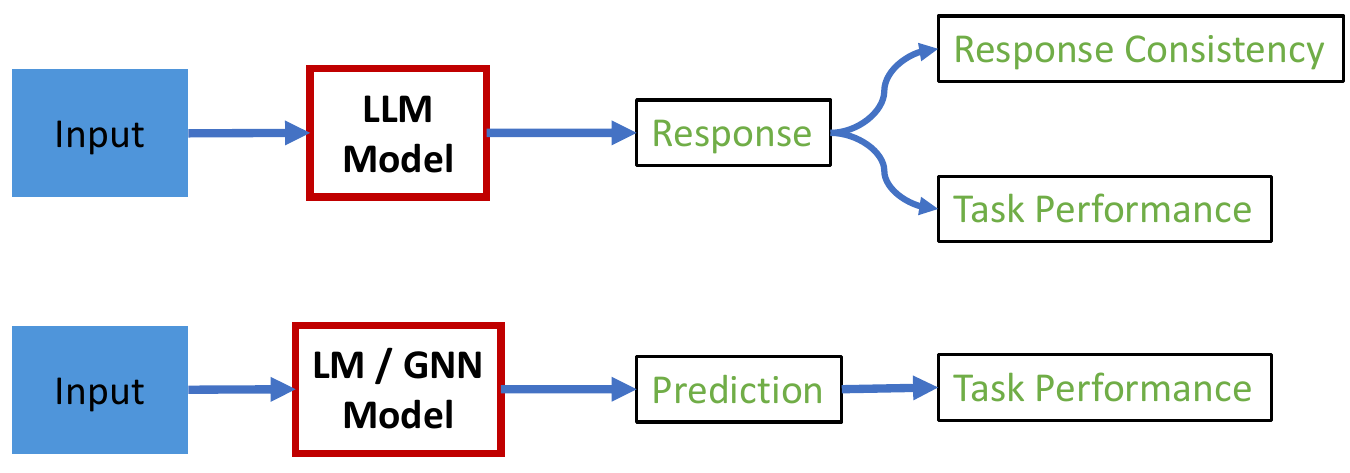}
\caption{
Overview of the evaluation process. 
}
\label{fig:evaluation_process}
\end{figure*}

\smallskip\noindent
\textbf{Evaluation Process and Setting.}
Our evaluation process workflow is illustrated in Figure~\ref{fig:evaluation_process}. 
In addition to the conventional evaluation workflow, which assesses the model's performance on downstream tasks, we also analyse the Language Model's (LLM) \emph{response consistency}. 
Given that LLMs may produce knowledge \emph{hallucinations}~\cite{HYMZ23}, generating responses that deviate from users' expectations, we calculate the ratio of LLM responses conforming to the required format, termed as response consistency.
To ensure fair comparisons, we adopt the fixed splits provided by \citet{HFRNDL21} in this paper. 
These splits guarantee consistency in evaluation conditions across different experiments, thereby facilitating meaningful comparisons between models.

\subsection{Performance}
\label{subsec:performance}

\begin{figure*}[!ht]
\centering
\includegraphics[width=.6\linewidth]{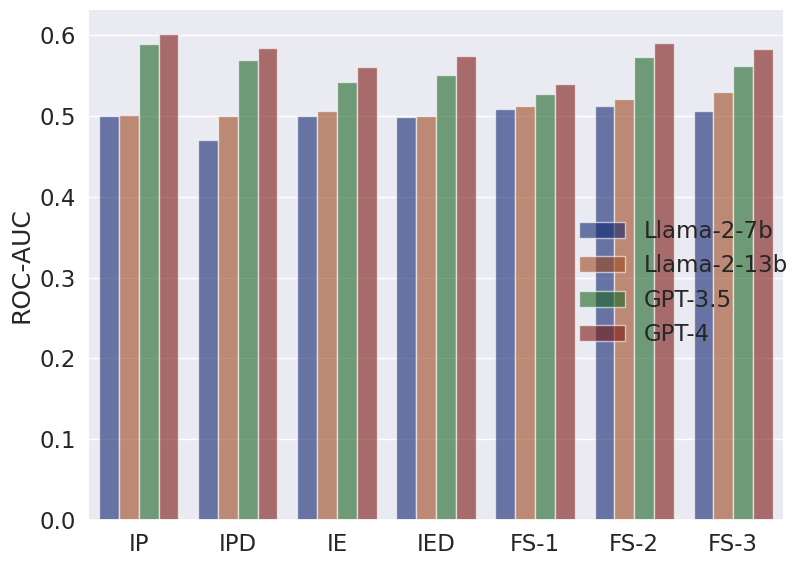}
\caption{
    Molecule graph property prediction performance of different LLMs for the \hiv dataset. 
}
\label{fig:diff_llms}
\end{figure*}

\smallskip\noindent
\textbf{Observation 1: GPT models consistently outperform other LLMs in handling molecule prediction tasks.}
In our initial investigations, we conduct a series of experiments to assess the efficacy of various LLMs on the \hiv dataset. 
We employ generated prompts following the templates outlined in Section~\ref{subsec:prompt_engineering}, encompassing \IP, \IPD, \IE, \IED, \FS-1, \FS-2, and \FS-3.
Figure~\ref{fig:diff_llms} presents our findings, demonstrating a consistent trend wherein GPT models exhibit superior performance compared to the Llama models across all evaluated metrics. 
This observation suggests that GPT models possess enhanced capabilities for handling molecule prediction tasks.
Furthermore, it's noteworthy that utilising the GPT-4 API incurs a cost that is 20 times higher than employing GPT-3.5. 
Additionally, our experiments reveal that the response time of the GPT-4 API is 10 times slower than that of GPT-3.5. 
Consequently, considering both performance and computational efficiency, we opt to employ GPT-3.5 as our default LLM for subsequent experiments in this study.

\smallskip\noindent
\textbf{Observation 2: LLMs are not potent experts for molecule prediction tasks.}
Analysing the performance of LLMs in molecule graph property prediction across six datasets as shown in Table~\ref{table:performance_llm_vs_sota}, it becomes apparent that LLMs consistently underperform compared to the three ML models. 
This trend suggests that relying on LLMs as experts for molecule prediction tasks may not yield satisfactory results. 
Further exploration is warranted to understand the limitations of LLMs in this domain and to explore alternative approaches for improving prediction accuracy.

\input{tables/table-performance-llm-vs-sota}

\begin{figure*}[!ht]
\centering
\includegraphics[width=1.\linewidth]{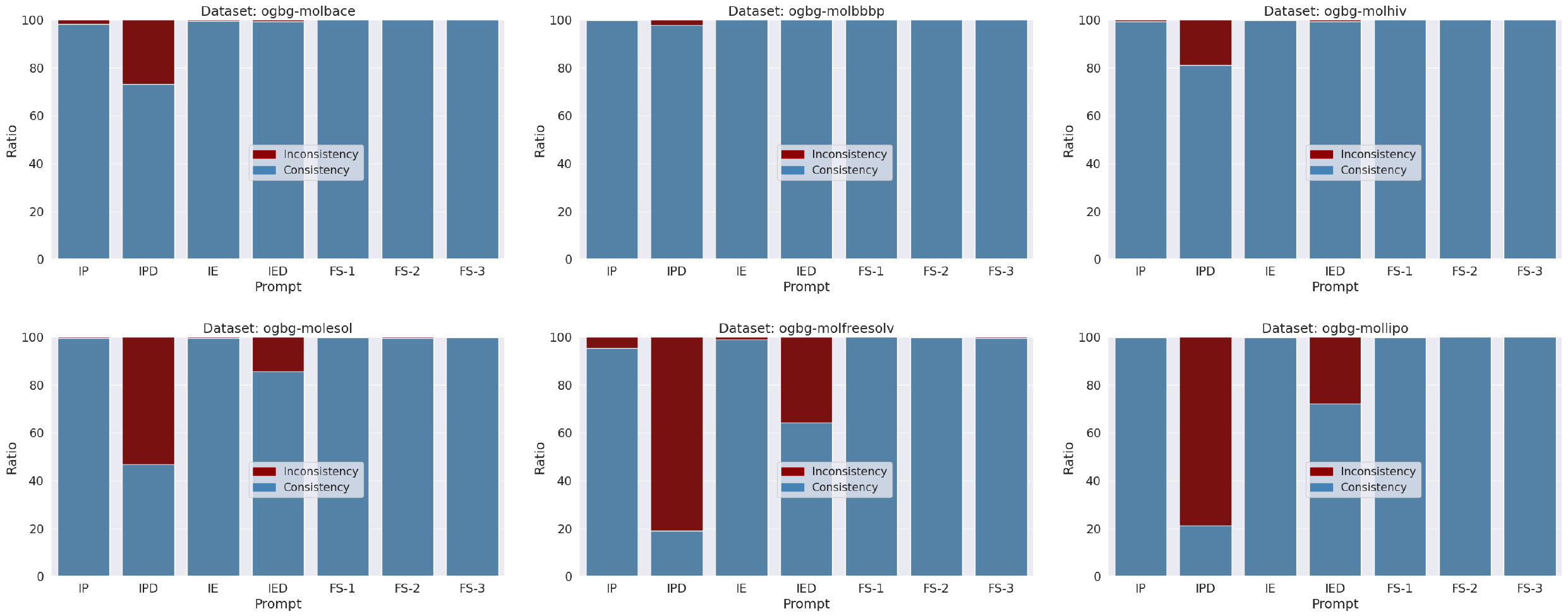}
\caption{
    Response consistency for the \bace, \bbbp, \hiv, \esol, \freesolv and \lipo datasets. 
}
\label{fig:consistency}
\end{figure*}

\smallskip\noindent
\textbf{Observation 3: Descriptions ($D$) do not help LLMs to understand molecule geometry structure.}
One significant constraint of current LLMs is their reliance on unstructured text, which limits their ability to incorporate crucial geometric structures inherent in molecules as input~\cite{LLWL23,GDL23}. 
To address this limitation, \citet{FHP23} propose encoding the graph structure into textual descriptions. 
Expanding on this concept, we integrate both atom features and graph structure into textual descriptions.
However, our findings from Table~\ref{table:performance_llm_vs_sota} reveal that augmenting prompts with descriptions does not consistently enhance performance; rather, it detrimentally affects performance in some instances. 
Furthermore, the decrease in response consistency reported in Figure~\ref{fig:consistency}, when descriptions are added to prompts, suggests that such additions may hinder LLMs' ability to maintain adherence to format requirements. 
We attribute this to the increased complexity introduced by the additional tokens in the description, thereby exacerbating the LLMs' attentional challenges.
Consequently, the practice of converting geometry structures into text for LLM consumption appears to be an insufficient solution. 
Addressing this limitation and exploring alternative strategies for effectively incorporating molecule geometry into LLMs remain promising areas for future investigation.

\smallskip\noindent
\textbf{Observation 4: The significance of geometric structure.}
The findings presented in Table~\ref{table:performance_llm_vs_sota} emphasise the superiority of models that integrate geometric structure compared to those that do not, underscoring the pivotal role of geometric information in precisely predicting a molecule's properties. 
Despite the evident importance of geometric structure, existing LLMs face constraints in directly incorporating this information into prompts, primarily due to limitations in token count within generated descriptions exceeding the LLM's constraints.
This limitation hampers the LLM's ability to effectively leverage geometric details, potentially compromising the accuracy of predictions. 
Thus, addressing this challenge represents a promising avenue for future research, with potential solutions including innovative token management techniques, refined prompt engineering strategies, or alternative model architectures capable of handling more extensive input representations. 
By overcoming this hurdle, LLMs can be empowered to better capture the intricacies of molecule geometry, thereby enhancing their predictive capabilities in various chemical modelling tasks.

\input{tables/table-performance-duo}

\input{tables/table-performance-trio}

\smallskip\noindent
\textbf{Observation 5: LLMs exhibit significant effectiveness as ML augmenters.}
In addition to utilising LLMs for direct molecule prediction tasks, we delve into the potential benefits of integrating LLMs with existing ML models. 
Following the framework delineated in Figure~\ref{fig:framework}, we augment the input features of ML models, specifically GNNs, with responses generated by LLMs.
The results presented in both Table~\ref{table:performance_duo} and Table~\ref{table:performance_trio} indicate a discernible enhancement in predictive performance upon introducing LLM responses as additional input features. 
This suggests that leveraging LLM-generated responses can effectively supplement the information captured by traditional ML models, leading to improved predictive accuracy.
Furthermore, established ML models, such as GNN variants, offer viable solutions for capturing geometric structure information inherent in molecules. 
By integrating LLM-generated responses with these models, we can potentially enhance their capacity to comprehend and leverage complex geometric features, thus further refining their predictive capabilities in chemical modelling tasks. This hybrid approach represents a promising avenue for advancing the state-of-the-art in molecular property prediction.

%% file: tables/table-dataset.tex
\begin{table*}[!ht]
\caption{
    Datasets statistics and splits from benchmark~\cite{WRFGGPLP18,HFZDRLCL20}. 
}
\label{table:dataset}
\centering
\begin{tabular}{l|r|r|r|r|r|r|r}
\hline
\hline
\textbf{Dataset} & \textbf{\#Graphs} & \textbf{\specialcell{Avg. \\ \#Nodes}}  & \textbf{\specialcell{Avg. \\ \#Edges}} &  \textbf{\#Train} & \textbf{\#Valid} & \textbf{\#Test} & \textbf{Task Type} \\
\hline
\bace~\cite{WRFGGPLP18} & 1,513 & 34.1 & 73.7 & 1,210 & 151 & 152 & Binary class. \\
\bbbp~\cite{WRFGGPLP18} & 2,039 & 24.1 & 51.9 & 1,631 & 204 & 204 & Binary class. \\
\hiv~\cite{WRFGGPLP18,HFZDRLCL20} & 41,127 & 25.5 & 27.5 & 32,901 & 4,113 & 4,113 & Binary class. \\
\esol~\cite{WRFGGPLP18} & 1,128 & 13.3 & 27.4 & 902 & 113 & 113 & Regression \\
\freesolv~\cite{WRFGGPLP18} & 642 & 8.7 & 16.8 & 513 & 64 & 65 & Regression \\
\lipo~\cite{WRFGGPLP18} & 4,200 & 27.0 & 59.0 & 3,360 & 420 & 420 & Regression \\
\hline
\end{tabular}
\end{table*}

%% file: tables/table-performance-llm-vs-sota.tex
\begin{table*}[!ht]
\caption{
    Molecule graph property prediction performance for the \bace, \bbbp, \hiv, \esol, \freesolv and \lipo datasets. 
    LM and GNN models follow the \solo pipeline. 
    Classification tasks are evaluated on ROC-AUC ($\uparrow$: higher is better), and regression tasks are evaluated on RMSE ($\downarrow$: lower is better). 
    The best test performance of LLM is marked with \underline{underline}. 
}
\label{table:performance_llm_vs_sota}
\centering
\resizebox{1.\linewidth}{!}{
\begin{tabular}{l|cc|cc|cc|cc|cc|cc}
\hline
\hline
 & \multicolumn{2}{c|}{\textbf{\bace}} & \multicolumn{2}{c|}{\textbf{\bbbp}} & \multicolumn{2}{c|}{\textbf{\hiv}} & \multicolumn{2}{c|}{\textbf{\esol}} & \multicolumn{2}{c|}{\textbf{\freesolv}} & \multicolumn{2}{c}{\textbf{\lipo}} \\
\hline
 & \multicolumn{6}{c|}{ROC-AUC $\uparrow$} & \multicolumn{6}{c}{RMSE $\downarrow$} \\
\hline
 & Valid & Test & Valid & Test & Valid & Test & Valid & Test & Valid & Test & Valid & Test \\
\hline
LLM$^{\IP}$     & \underline{0.5690} & 0.5756
                & 0.4606 & \underline{0.5399}
                & 0.5519 & \underline{0.5892}
                & 2.6221 & 2.0422
                & 6.1699 & \underline{4.4421}
                & 1.9836 & 1.8411 \\
LLM$^{\IPD}$    & 0.4835 & 0.5534
                & 0.4643 & 0.4664
                & 0.4732 & 0.5693
                & 3.7395 & 3.1721
                & 8.1598 & 7.2877
                & 2.6464 & 2.5046 \\
LLM$^{\IE}$     & 0.4769 & 0.5220
                & 0.4463 & 0.5237
                & 0.5487 & 0.5419
                & 2.1055 & 2.5549
                & 5.9059 & 4.3097
                & 2.1044 & 1.9158 \\
LLM$^{\IED}$    & 0.5299 & 0.4761
                & 0.4742 & 0.4091
                & 0.5361 & 0.5512
                & 3.9001 & 4.2289
                & 7.4837 & 5.3689
                & 2.4191 & 2.4219 \\
LLM$^{\FS-1}$   & 0.4822 & 0.5122
                & 0.5955 & 0.4954
                & 0.5229 & 0.5268
                & \underline{1.7699} & 2.8762
                & 6.4785 & 4.7553
                & 1.9810 & 1.8432 \\
LLM$^{\FS-2}$   & 0.4277 & \underline{0.6090}
                & \underline{0.6019} & 0.5075
                & \underline{0.5619} & 0.5731
                & 1.9271 & 2.1020
                & \underline{5.5078} & 4.5606
                & 1.9138 & 1.8118 \\
LLM$^{\FS-3}$   & 0.5405 & 0.5949
                & 0.6000 & 0.5388
                & 0.5475 & 0.5616
                & 1.9548 & \underline{1.9963}
                & 6.3753 & 4.7241
                & \underline{1.8291} & \underline{1.7923} \\
\hline
LM              & 0.5584 & 0.6163
                & 0.9307 & 0.6727
                & 0.5305 & 0.5037
                & 2.1139 & 2.2549
                & 6.6189 & 4.4532
                & 1.2095 & 1.1066 \\
GCN             & 0.7879 & 0.7147
                & 0.9582 & 0.6707
                & 0.8461 & 0.7376
                & 0.8538 & 1.2561
                & 2.8275 & 2.5096
                & 0.6985 & 0.7201 \\
GIN             & 0.8012 & 0.7673
                & 0.9608 & 0.6708
                & 0.8406 & 0.7601
                & 0.8010 & 0.9555
                & 2.2106 & 2.1610
                & 0.6482 & 0.7019 \\
\hline
\end{tabular}
}
\end{table*}

%% file: tables/table-performance-duo.tex
\begin{table*}[!ht]
\caption{
    Molecule graph property prediction performance for the \bace, \bbbp, \hiv, \esol, \freesolv and \lipo datasets, follow the \textsc{Duo} pipeline. 
    Classification tasks are evaluated on ROC-AUC ($\uparrow$: higher is better), and regression tasks are evaluated on RMSE ($\downarrow$: lower is better). 
    The best performance of each model is marked with \underline{underline}, and the overall best performance is marked as \textbf{bold}. 
}
\label{table:performance_duo}
\centering
\resizebox{1.\linewidth}{!}{
\begin{tabular}{l|cc|cc|cc|cc|cc|cc}
\hline
\hline
 & \multicolumn{2}{c|}{\textbf{\bace}} & \multicolumn{2}{c|}{\textbf{\bbbp}} & \multicolumn{2}{c|}{\textbf{\hiv}} & \multicolumn{2}{c|}{\textbf{\esol}} & \multicolumn{2}{c|}{\textbf{\freesolv}} & \multicolumn{2}{c}{\textbf{\lipo}} \\
\hline
 & \multicolumn{6}{c|}{ROC-AUC $\uparrow$} & \multicolumn{6}{c}{RMSE $\downarrow$} \\
\hline
 & Valid & Test & Valid & Test & Valid & Test & Valid & Test & Valid & Test & Valid & Test \\
\hline
LM              & 0.5584 & 0.6163
                & 0.9307 & 0.6727
                & 0.5305 & 0.5037
                & 2.1139 & 2.2549
                & \underline{6.6189} & 4.4532
                & \underline{1.2095} & 1.1066 \\
LM$^{\IF}$      & 0.6075 & 0.6045
                & 0.9347 & 0.6664
                & 0.5669 & 0.5453
                & 2.1292 & 2.2687
                & 6.6526 & 4.4754
                & 1.2112 & 1.1057 \\
LM$^{\IFD}$     & 0.5357 & 0.6012
                & 0.9401 & 0.6566
                & 0.5323 & 0.5209
                & 2.1193 & 2.2614
                & 6.6639 & 4.4974
                & 1.2096 & 1.1026 \\
LM$^{\IE}$      & 0.5851 & 0.6107
                & \underline{0.9404} & 0.6694
                & \underline{0.5760} & 0.5514
                & 2.1264 & 2.2687
                & 6.6657 & 4.4989
                & 1.2109 & 1.1068 \\
LM$^{\IED}$     & 0.6228 & 0.6107
                & 0.9370 & 0.6735
                & 0.5487 & 0.5342
                & 2.1158 & 2.2587
                & 6.6737 & 4.5062
                & 1.2085 & \underline{1.1020} \\
LM$^{\FS-1}$    & 0.6059 & 0.6034
                & 0.9401 & 0.6576
                & 0.5458 & 0.5037
                & 2.1319 & 2.2729
                & 6.6516 & 4.5003
                & 1.2103 & 1.1039 \\
LM$^{\FS-2}$    & \underline{0.6648} & \underline{0.6287}
                & 0.9298 & \underline{0.6738}
                & 0.5458 & 0.5037
                & \underline{2.1096} & \underline{2.2519}
                & 6.6772 & 4.5148
                & 1.2093 & 1.1040 \\
LM$^{\FS-3}$    & 0.5937 & 0.6165
                & 0.9299 & 0.6645
                & 0.5909 & \underline{0.5875}
                & 2.1194 & 2.2597
                & 6.6326 & \underline{4.4386}
                & 1.2104 & 1.1063 \\
\hline
GCN             & 0.7879 & 0.7147
                & \underline{0.9582} & 0.6707
                & 0.8461 & 0.7376
                & 0.8538 & 1.2561
                & 2.8275 & 2.5096
                & 0.6985 & 0.7201 \\
GCN$^{\SMILES}$ & 0.7620 & 0.7107
                & 0.9572 & 0.6688
                & 0.8483 & 0.7306
                & 0.8495 & 0.9537
                & 2.8358 & 2.4803
                & \underline{0.6843} & 0.7132 \\
GCN$^{\IF}$     & 0.7855 & 0.7542
                & 0.9575 & 0.6647
                & \textbf{\underline{0.8602}} & 0.7535
                & 0.8571 & 0.9534
                & 2.8351 & 2.2911
                & 0.6897 & 0.7037 \\
GCN$^{\IFD}$    & 0.8004 & 0.7581
                & 0.9543 & 0.6710
                & 0.8572 & 0.7416
                & 0.8457 & 1.1430
                & 2.7641 & \underline{2.2033}
                & 0.6939 & 0.7187 \\
GCN$^{\IE}$     & \underline{0.8017} & 0.7396
                & 0.9577 & \underline{0.6741}
                & 0.8491 & 0.7543
                & 0.8439 & 0.9559
                & \underline{2.6150} & 2.2197
                & 0.6846 & 0.6962 \\
GCN$^{\IED}$    & 0.7936 & 0.7020
                & 0.9562 & 0.6550
                & 0.8497 & \underline{0.7573}
                & \underline{0.8369} & 0.9650
                & 2.7071 & 2.2775
                & 0.7100 & \underline{0.6953} \\
GCN$^{\FS-1}$   & 0.7933 & 0.7535
                & 0.9558 & 0.6689
                & 0.8446 & 0.7412
                & 0.8569 & 0.9439
                & 2.7189 & 2.4021
                & 0.6984 & 0.6991 \\
GCN$^{\FS-2}$   & 0.7905 & \textbf{\underline{0.7903}}
                & 0.9566 & 0.6704
                & 0.8529 & 0.7480
                & 0.8557 & \underline{0.9402}
                & 2.7440 & 2.3545
                & 0.6873 & 0.7245 \\
GCN$^{\FS-3}$   & 0.7933 & 0.7682
                & 0.9566 & 0.6552
                & 0.8520 & 0.7544
                & 0.8567 & 0.9441
                & 2.7580 & 2.3575
                & 0.7066 & 0.7213 \\
\hline
GIN             & 0.8012 & 0.7673
                & \textbf{\underline{0.9608}} & 0.6708
                & 0.8406 & 0.7601
                & 0.8010 & 0.9555
                & \textbf{\underline{2.2106}} & 2.1610
                & 0.6482 & 0.7019 \\
GIN$^{\SMILES}$ & 0.7975 & 0.7802
                & 0.9579 & 0.6597
                & \underline{0.8548} & \textbf{\underline{0.7837}}
                & 0.8174 & \textbf{\underline{0.9142}}
                & 2.6264 & 2.3493
                & 0.6440 & 0.6920 \\
GIN$^{\IF}$     & 0.7976 & \underline{0.7804}
                & 0.9583 & \textbf{\underline{0.6773}}
                & 0.8497 & 0.7684
                & \textbf{\underline{0.7797}} & 0.9349
                & 2.4530 & 2.2394
                & 0.6454 & 0.6966 \\
GIN$^{\IFD}$    & 0.8001 & 0.7593
                & 0.9577 & 0.6763
                & 0.8431 & 0.7577
                & 0.7843 & 0.9497
                & 2.5720 & 2.1906
                & 0.6507 & \textbf{\underline{0.6836}} \\
GIN$^{\IE}$     & 0.8046 & 0.7549
                & 0.9606 & 0.6762
                & 0.8481 & 0.7339
                & 0.7924 & 0.9617
                & 2.3844 & \textbf{\underline{2.1509}}
                & 0.6384 & 0.6909 \\
GIN$^{\IED}$    & 0.7916 & 0.7589
                & 0.9604 & 0.6494
                & 0.8493 & 0.7798
                & 0.7961 & 0.9623
                & 2.4639 & 2.1956
                & 0.6424 & 0.6846 \\
GIN$^{\FS-1}$   & 0.8001 & 0.7594
                & 0.9603 & 0.6748
                & 0.8547 & 0.7504
                & 0.7863 & 0.9661
                & 2.4021 & 2.2439
                & 0.6422 & 0.6950 \\
GIN$^{\FS-2}$   & \textbf{\underline{0.8062}} & 0.7677
                & 0.9595 & 0.6659
                & 0.8492 & 0.7824
                & 0.7949 & 0.9171
                & 2.3584 & 2.2675
                & \textbf{\underline{0.6399}} & 0.6969 \\
GIN$^{\FS-3}$   & 0.8038 & 0.7388
                & 0.9586 & 0.6762
                & 0.8515 & 0.7602
                & 0.8118 & 0.9155
                & 2.3942 & 2.3633
                & 0.6447 & 0.6865 \\
\hline
\end{tabular}
}
\end{table*}

%% file: tables/table-performance-trio.tex
\begin{table*}[!ht]
\caption{
    Molecule graph property prediction performance for the \bace, \bbbp, \hiv, \esol, \freesolv and \lipo datasets, follow the \textsc{Trio} pipeline. 
    Classification tasks are evaluated on ROC-AUC ($\uparrow$: higher is better), and regression tasks are evaluated on RMSE ($\downarrow$: lower is better). 
    GNN$^{\duo}$ indicates the best performance collected from Table~\ref{table:performance_duo}. 
    The best performance of each model is marked with \underline{underline}, and the overall best performance is marked as \textbf{bold}. 
}
\label{table:performance_trio}
\centering
\resizebox{1.\linewidth}{!}{
\begin{tabular}{l|cc|cc|cc|cc|cc|cc}
\hline
\hline
 & \multicolumn{2}{c|}{\textbf{\bace}} & \multicolumn{2}{c|}{\textbf{\bbbp}} & \multicolumn{2}{c|}{\textbf{\hiv}} & \multicolumn{2}{c|}{\textbf{\esol}} & \multicolumn{2}{c|}{\textbf{\freesolv}} & \multicolumn{2}{c}{\textbf{\lipo}} \\
\hline
 & \multicolumn{6}{c|}{ROC-AUC $\uparrow$} & \multicolumn{6}{c}{RMSE $\downarrow$} \\
\hline
 & Valid & Test & Valid & Test & Valid & Test & Valid & Test & Valid & Test & Valid & Test \\
\hline
GCN             & 0.7879 & 0.7147
                & 0.9582 & 0.6707
                & 0.8461 & 0.7376
                & 0.8538 & 1.2561
                & 2.8275 & 2.5096
                & 0.6985 & 0.7201 \\
GCN$^{\duo}$    & \underline{0.8017} & 0.7903
                & 0.9582 & 0.6741
                & \textbf{\underline{0.8602}} & 0.7573
                & 0.8369 & 0.9402
                & \underline{2.6150} & \underline{2.2033}
                & \underline{0.6843} & 0.6953 \\
GCN$^{\SAIF}$   & 0.7886 & 0.7566
                & 0.9602 & 0.6564
                & 0.8430 & \underline{0.7649}
                & 0.8461 & \underline{0.9366}
                & 2.7679 & 2.4457
                & 0.6900 & 0.7040 \\
GCN$^{\SAIFD}$  & 0.7900 & 0.7896
                & 0.9540 & 0.6700
                & 0.8322 & 0.7539
                & 0.8379 & 9.6060
                & 2.8279 & 2.3047
                & 0.7020 & 0.7105 \\
GCN$^{\SAIE}$   & 0.8044 & 0.7727
                & 0.9579 & 0.6706
                & 0.8548 & 0.7643
                & \underline{0.8312} & 1.4489
                & 2.7831 & 2.2401
                & 0.6921 & 0.7170 \\
GCN$^{\SAIED}$  & 0.7811 & 0.7587
                & 0.9576 & 0.6639
                & 0.8470 & 0.7414
                & 0.8560 & 0.9598
                & 2.7709 & 2.2807
                & 0.6975 & 0.7256 \\
GCN$^{\SAFS-1}$ & 0.7899 & 0.7662
                & 0.9578 & \underline{0.6774}
                & 0.8529 & 0.7584
                & 0.8548 & 0.9419
                & 2.7675 & 2.4254
                & 0.7051 & 0.7101 \\
GCN$^{\SAFS-2}$ & 0.7851 & 0.7625
                & \textbf{\underline{0.9624}} & 0.6679
                & 0.8592 & 0.7386
                & 0.8496 & 1.9680
                & 2.7437 & 2.3693
                & 0.7015 & 0.7015 \\
GCN$^{\SAFS-3}$ & 0.7950 & \underline{\underline{0.7939}}
                & 0.9593 & 0.6710
                & 0.8471 & 0.7635
                & 0.8810 & 1.0457
                & 2.6897 & 2.2867
                & 0.6939 & \underline{0.6937} \\
\hline
GIN             & 0.8012 & 0.7673
                & 0.9608 & 0.6708
                & 0.8406 & 0.7601
                & 0.8010 & 0.9555
                & 2.2106 & 2.1610
                & 0.6482 & 0.7019 \\
GIN$^{\duo}$    & \textbf{\underline{0.8062}} & 0.7804
                & \underline{0.9608} & 0.6773 
                & \underline{0.8548} & 0.7837
                & \textbf{\underline{0.7797}} & 0.9142
                & \textbf{\underline{2.2106}} & 2.1509
                & 0.6399 & 0.6836  \\
GIN$^{\SAIF}$   & 0.8026 & 0.7606
                & 0.9547 & 0.6728
                & 0.8345 & 0.7667
                & 0.7900 & 0.9547
                & 2.4822 & 2.1688
                & \textbf{\underline{0.6388}} & \textbf{\underline{0.6791}} \\
GIN$^{\SAIFD}$  & 0.7974 & \textbf{\underline{0.7996}}
                & 0.9584 & 0.6751
                & 0.8351 & 0.7549
                & 0.7956 & 0.9305
                & 2.3015 & 2.2223
                & 0.6465 & 0.6891 \\
GIN$^{\SAIE}$   & 0.8035 & 0.7459
                & 0.9597 & 0.6719
                & 0.8475 & 0.7591
                & 0.8217 & 0.9320
                & 2.3971 & 2.2266
                & 0.6423 & 0.6859 \\
GIN$^{\SAIED}$  & 0.7914 & 0.7638
                & 0.9594 & 0.6584
                & 0.8519 & 0.7512
                & 0.7845 & 0.9599
                & 2.3271 & 2.1759
                & 0.6454 & 0.6900 \\
GIN$^{\SAFS-1}$ & 0.7989 & 0.7542
                & 0.9591 & 0.6556
                & 0.8400 & \textbf{\underline{0.7922}}
                & 0.8010 & \textbf{\underline{0.9130}}
                & 2.4698 & 2.2435
                & 0.6490 & 0.6967 \\
GIN$^{\SAFS-2}$ & 0.7886 & 0.7704
                & 0.9604 & \textbf{\underline{0.6789}}
                & 0.8461 & 0.7619
                & 0.7967 & 0.9423
                & 2.4492 & \textbf{\underline{2.1307}}
                & 0.6452 & 0.6939 \\
GIN$^{\SAFS-3}$ & 0.8013 & 0.7762
                & 0.9587 & 0.6732
                & 0.8493 & 0.7645
                & 0.8099 & 0.9302
                & 2.3327 & 2.1829
                & 0.6407 & 0.6912 \\
\hline
\end{tabular}
}
\end{table*}

%% file: pages/conclusion.tex
In conclusion, our empirical investigations have provided valuable insights into the capacity of LLMs to handle molecular tasks. Our comprehensive analysis across six benchmark datasets has demonstrated that LLMs generally exhibit less competitive performance in molecular prediction tasks when compared to established ML models, especially those explicitly designed to capture geometric structures within molecules.
Moreover, our findings underscore the potential of leveraging LLMs as complementary tools to enhance the performance of existing ML models. By integrating LLMs as augmenters, we observed improvements in predictive accuracy, suggesting a promising avenue for effectively harnessing the capabilities of both LLMs and traditional ML models in tandem.
While our study highlights LLMs' current limitations in molecular tasks, it also opens up new avenues for future research. Exploring innovative methodologies to better integrate LLMs with domain-specific knowledge and structural information could potentially bridge the performance gap observed in our experiments.
Overall, our work contributes to a deeper understanding of the strengths and weaknesses of LLMs in molecular tasks, paving the way for more informed strategies for leveraging these models for practical applications in chemistry, biology, and related fields.

Our work makes an important step toward leveraging LLMs for challenging molecule prediction tasks. 
There exist numerous promising avenues for future work. 
One critical area for further investigation is addressing the inherent limitation of LLMs in comprehending molecule geometric structures. 
As evidenced by the performance disparities highlighted in Section~\ref{subsec:performance}, the inability of LLMs to grasp such structural nuances often results in inaccuracies. 
Therefore, overcoming this limitation and enhancing LLMs' understanding of molecule geometric structure is imperative for their broader applicability in molecular tasks.
Additionally, while our study introduces straightforward yet effective frameworks for integrating LLMs with traditional ML models, there remains ample room for developing more sophisticated methodologies in this regard. 
Designing advanced frameworks that seamlessly incorporate LLMs with existing ML models presents a promising avenue for future research, potentially yielding enhanced predictive performance and model interpretability~\cite{ZBM23}.
Finally, developing \emph{molecule specialist LLMs} would be impressive to the community. 
Despite LLMs' underperformance compared to baselines across many tasks, their ability to derive solutions from limited examples underscores their potential for generalised intelligence in the molecular domain. 
However, current LLMs exhibit notable hallucinations in chemistry tasks, suggesting room for improvement. 
Continuous LLM development and research into mitigating hallucinations offer optimism for enhancing their efficacy in practical chemistry problem-solving.